\documentclass[11pt]{article}
\usepackage{a4wide,parskip,appendix}
\usepackage{amsfonts}
\usepackage{pdfpages}
\usepackage{mdframed}
\usepackage{titlesec}
\usepackage{venndiagram}

\usepackage{amsmath}
\DeclareMathOperator*{\argmax}{arg\,max}
\DeclareMathOperator*{\argmin}{arg\,min}

\usepackage{hyperref}
\usepackage{caption}
\usepackage{algorithm2e}

\usepackage{xcolor}
\definecolor{LRcolour}{RGB}{200,250,200}
\definecolor{Dcolour}{RGB}{210,230,250}
\definecolor{MixColour}{RGB}{250,230,200}

\usepackage{csquotes}
\MakeOuterQuote{"}

\usepackage[a4paper, total={6in, 9in}]{geometry}

\titlespacing*{\section}
{0pt}{5.5ex plus 1ex minus .2ex}{4.3ex plus .2ex}
\titlespacing*{\subsection}
{0pt}{5.5ex plus 1ex minus .2ex}{2.3ex plus .2ex}

\newenvironment{aside}
  {\begin{mdframed}[style=0,%
      leftline=false,rightline=false,leftmargin=2em,rightmargin=2em,%
          innerleftmargin=0pt,innerrightmargin=0pt,linewidth=0.75pt,%
      skipabove=12pt,skipbelow=12pt,fontcolor=darkgray]\small}
  {\end{mdframed}}

\newcommand{\asidetitle}[1]{\large{\textit{#1}}\small}

\newcommand{\abs}[1]{\lVert#1\rVert}
\RestyleAlgo{ruled}

\newcommand{\textfn}[1]{\textsc{\small{#1}}}

\newcommand{\highlight}[2][yellow]{\mathchoice%
  {\colorbox{#1}{$\displaystyle#2$}}%
  {\colorbox{#1}{$\textstyle#2$}}%
  {\colorbox{#1}{$\scriptstyle#2$}}%
  {\colorbox{#1}{$\scriptscriptstyle#2$}}}%

\begin{document}

\title{\textbf{Learning-Rate-Free Learning: Dissecting D-Adaptation and Probabilistic Line Search}}
\author{Max McGuinness (mgm52), Queens' College \\ \emph{R255 Topic 4 - Probabilistic Numerics}}
\maketitle

\begin{abstract}
This paper explores two recent methods for learning rate optimisation in stochastic gradient descent: \textit{D-Adaptation} \cite{defazio2023dadaptation} and \textit{probabilistic line search} \cite{mahsereci2015probabilisticlinesearch}. These approaches aim to alleviate the burden of selecting an initial learning rate by incorporating distance metrics and Gaussian process posterior estimates, respectively. In this report, I provide an intuitive overview of both methods, discuss their shared design goals, and devise scope for merging the two algorithms.

\tableofcontents
\newpage

\end{abstract}

\section{Introduction}
This report investigates the problem of learning rate optimisation, focusing on techniques that remove the programmer's burden to choose a proper initial learning rate.

The report aims to satisfy two purposes:
\begin{enumerate}
    \item \textbf{Acting as an intuition-led guide} to Defazio and Mishchenko's 2023 \textit{Learning-Rate-Free Learning by D-Adaptation} \cite{defazio2023dadaptation} and Mahsereci and Hennig's 2015 \textit{Probabilistic Line Searches for Stochastic Optimisation} \cite{mahsereci2015probabilisticlinesearch}.
    \item \textbf{Presenting a unified notation} to discuss optimisation techniques, allowing us to bring together the two learning-rate-free approaches and introduce probabilistics to \textit{D-Adaptation} in the Discussion section (\ref{sec:discussion}).
\end{enumerate}

We will begin by recapping the general problem of optimisation. This will establish a common language through which to discuss optimisation algorithms, and introduce the notation used in Defazio et al's D-Adaptation paper.

\subsection{Optimisation Problems}
Given an arbitrary function $f : \mathbb{R}^p \rightarrow \mathbb{R}$, one can define the problem of optimisation\footnote{Global, continuous optimisation.} as identifying any solution $x_*$ such that:
\begin{equation} \label{eq:nonstoch_minima}
    x_* \in \argmin_{x \in \mathbb{R}^p} f(x).
\end{equation}

This problem can be further categorised by properties of $f$, described in the following subsections.

\subsubsection{Convex/Non-Convex Optimisation} \label{sec:convex_opt}
Convex optimisation is concerned with the minimisation of a \textbf{convex} function $f$, which enables guarantees of global convergence for simple methods.

\begin{aside}
\asidetitle{Aside: Convexity}

We can describe a continuous function as \textit{convex} when the values of the function between any two points on its graph lie beneath the line segment connecting those two points.

To put it formally:
\begin{align} \label{eq:f_convex}
    & f \; \text{is convex} \iff \notag \\
    & \forall \theta \in [0, 1]; x \in \mathbb{R}^p; x' \in \mathbb{R}^p : \notag \\
    & f(\theta x + (1 - \theta)x') \; \leq \; \theta f(x) + (1 - \theta) f(x').
\end{align}

This notion can be extended to \textbf{sets} of values by the intuitive notion that one should be able to interpolate between any two points in the set without leaving said set:
\begin{align}
    & C \; \text{is convex} \iff \notag \\
    & \forall \theta \in [0, 1]; x \in C; x' \in C : \notag \\
    & \theta x + (1 - \theta)x' \in C.
\end{align}

These two definitions can be consolidated by considering the epigraph of $f$,
\begin{equation}
    \text{epi}(f) = \{(x, y) | f(x) \leq y\},
\end{equation}
giving us the equivalence:
\begin{equation}
    \text{Function $f$ is convex $\iff$ epi$(f)$ is convex.}
\end{equation}
\end{aside}

\textbf{Gradient descent} (GD) is the archetypal means of solving convex minimisation for differentiable functions. At each timestamp $k$, GD translates $x$ against $f$'s gradient $g_k = \nabla f(x_k)$, in proportion to some step size $\gamma_k$:
\begin{equation} \label{eq:GD}
    x_{k+1} = x_k - \gamma_k g_k.
\end{equation}

One can apply gradient descent to \textbf{non-convex minimisation}, but this sacrifices significant theoretical guarantees: the algorithm can now get "stuck" in local minima, making global convergence far more challenging or even impossible to determine.

However, convex approaches remain appealing even in non-convex scenarios, as many non-convex optimisation problems resemble perturbed convex ones \cite{danilova2022nonconvex-sometimes-is-convexy}. That is to say, the optimal solution still exists in some parabola between -$\infty$ and $\infty$, with near-optimal local minima located nearby.

\subsubsection{Stochastic/Deterministic Optimisation} \label{sec:stochastic_opt}
\textbf{Stochastic optimisation} problems may involve random variables at any point in their formulation. For the purposes of this report, we will focus on problems in which this manifests as a random variable $\xi$ in the objective function $f$, such that the optimisation becomes:
\begin{equation} \label{eq:stoch_minima}
    x_* \in \argmin_{x \in \mathbb{R}^p} \mathop{\mathbb{E}} [f(x, \xi)].
\end{equation}

Impressively, \textbf{stochastic gradient descent} (SGD) has been shown to still converge to a local minimum in expected squared
error for diminishing learning rates \cite{robbins1951SGD, leon1998SGD-converges-to-local-min}.
 
 In addition, this stochasticity can actually be utilised positively for the \textit{non-convex global gradient descent} scenario mentioned in \ref{sec:convex_opt}, as:
 \begin{itemize}
     \item Its randomness (combined with careful choice of learning rate) can "nudge" the algorithm out of stuck positions in local minima. 
     \item In the context of mini-batch machine learning, SGD has been found to have an implicit \textit{regularising} effect that---by penalising the mean squared norm of each $g_k$---encourages generalisation \cite{smith2021SGD-implicit-regularization}.
 \end{itemize}

\subsubsection{Differentiable/Non-Differentiable Optimisation}
\textbf{Non-differentiable optimisation} is a subcategory of non-convex optimisation that deals with functions which have non-differentiable components, such as singularities or corners (e.g. consider the \texttt{abs} function).

One popular approach to non-differentiable optimisation is to use a \textbf{subgradient method}, which generalises gradient-based methods such as GD (equation \ref{eq:GD}) to non-differentiable functions.

\begin{aside}
\asidetitle{Aside: Subdifferentials}

Informally, a \textbf{subgradient} of $f$ at $x$ is a vector defining a hyperplane that intersects $f(x)$ yet underestimates $f$, such as in Figure \ref{fig:subgradient}.

\begin{center}
\includegraphics[width=5cm]{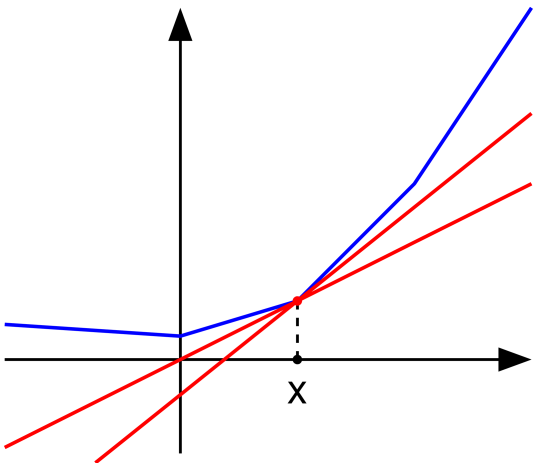} 
\captionsetup{width=0.8\linewidth}
\captionof{figure}{\small Two possible subgradients (red) at position \(x\) for some function \(f\) (blue).\protect\footnotemark}
\label{fig:subgradient}
\end{center}

The \textbf{subdifferential} $\partial f(x)$ denotes the set of all subgradients $g$ of $f$ at $x$, of which there are either 0 (\textit{if f is non-continuous}), 1 (\textit{if f is differentiable at x}), or infinity. Formally, subdifferential membership can be defined as:
\begin{align} \label{eq:subgradients}
    & g \in \partial f(x) \iff \notag \\
    & \forall x' \in \mathbb{R}^p : \notag \\
    & f(x') - f(x) \geq g \cdot (x' - x).
\end{align}

Usefully for optimisation, the subdifferential $\partial f(x)$ will always form a \textit{closed, convex} set. This means that one can use standard convex optimisation methods to identify, say, the presence of 0 in a subdifferential (motivated by $x=x_* \iff 0 \in \partial f(x)$ per equation \ref{eq:nonstoch_minima}).
\end{aside}
\footnotetext{Image adapted from \url{https://en.wikipedia.org/wiki/Subderivative\#/media/File:Subderivative\_illustration.png}}

In the case of gradient descent, we can generalise the method to non-differentiable functions by redefining gradient $g_k$ into the subgradient:
\begin{equation} \label{eq:subgrad-method}
    g_k \in \partial f(x_k).
\end{equation}

\subsection{Learning Rate Optimisation}
The problem of \textbf{learning rate optimisation} (LRO) is concerned with identifying the optimal learning rate (aka \textit{step size}) for iterative optimisation algorithms such as GD (equation \ref{eq:GD}). As approaches to LRO can take many forms, to encapsulate them it is easiest to begin talking in terms of time and computation.

Suppose $\Gamma = \{\gamma_0, ..., \gamma_n\}$ is the learning rate scheme used by our GD-like optimiser, and $\texttt{time}_{x_k}(\Gamma)$ the time at which this optimiser reaches some iterate $x_k$. We can define the general LRO problem to be finding a time-optimal scheme $\Gamma_*$ such that:
\begin{equation} \label{eq:LRO_problem}
     \Gamma_* \in \argmin_{\Gamma \in \mathbb{R}^n}(\texttt{time}_{x_*}(\Gamma)),
\end{equation}
where $x_*$ is the inner optimiser's goal per equation \ref{eq:nonstoch_minima} (or per some other stopping criteria, such as Wolfe conditions \cite{wolfe1959wolfeconditions} described in eq \ref{eq:wolfe1}, \ref{eq:wolfe2}).

Thus we can see LRO as an optimisation problem itself, but with the added complexity that evaluating each scheme $\Gamma$ is extremely computationally intensive for non-trivial choices of $f$. Furthermore, $\texttt{time}_{x_*}$ is almost always non-convex. This leads LRO methods to be highly inexact.

\subsubsection{Approaches to LRO}

With the above in mind, I would (very) broadly place learning rate optimisers into the following categories:

\begin{itemize}
    \item \textbf{Inter-step} - varies $\gamma$ once per step $x_k$.
    \item \textbf{Intra-step} - varies $\gamma$ many times per step $x_k$.
    \item \textbf{Inter-run} - varies $\gamma$ once per \textit{re-initialisation} of $x_k$ to $x_0$.
\end{itemize}

Additionally, I identify \textbf{dynamic} approaches (schemes that depend on mid-run variables, as opposed to instantiating a static $\Gamma$ at the start of the run) in \textbf{\underline{underline}}, summarised in Figure \ref{fig:LRO_venn}.

\begin{center}
\includegraphics[width=12cm]{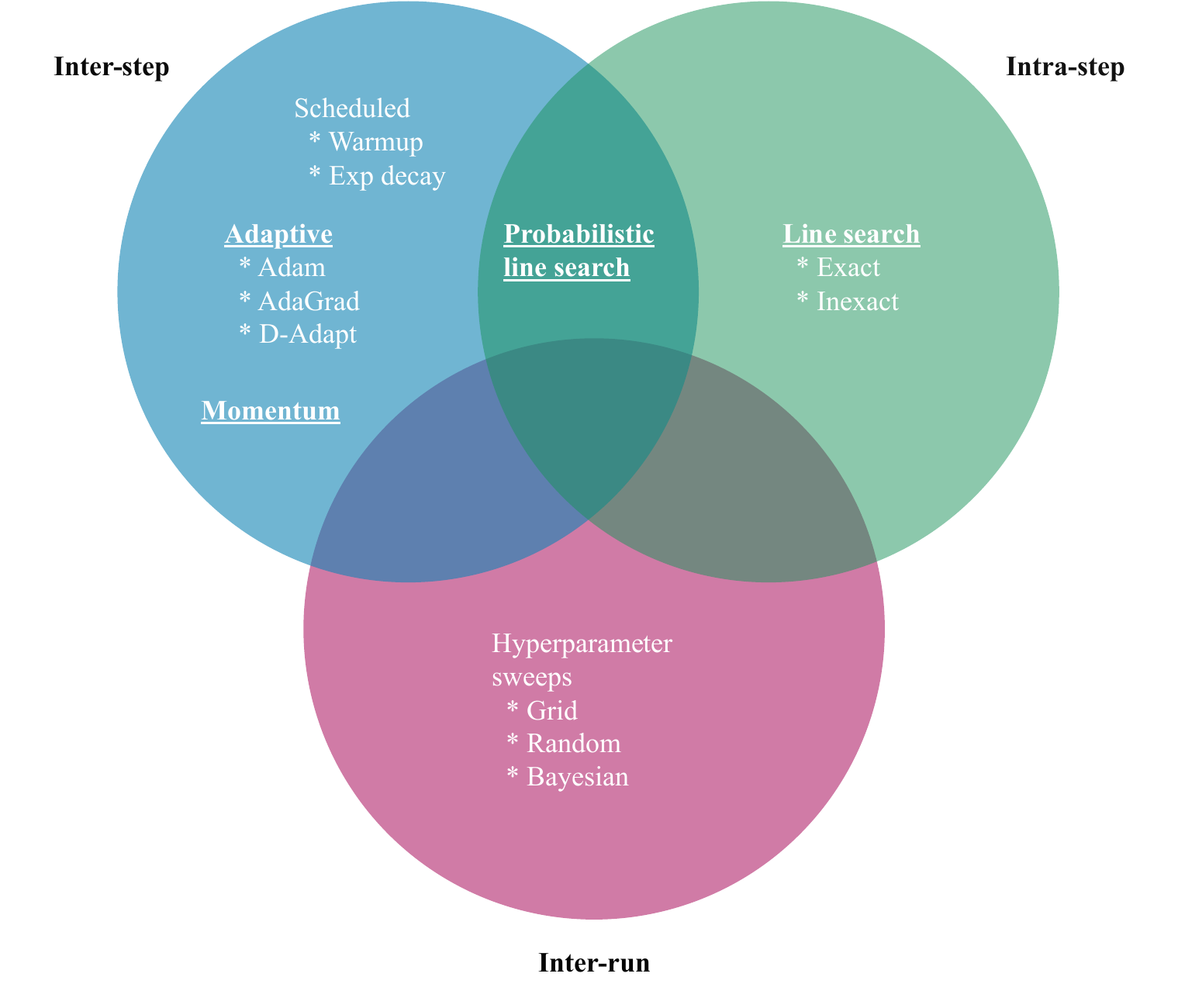} 
\captionsetup{width=1\linewidth}
\captionof{figure}{A categorisation of \textit{learning rate optimisation} approaches. Dynamic methods are indicated in \textbf{\underline{underline}}. Almost all of these techniques can be combined; consider finding Adam's hyperparameters via a sweep, for example.}
\label{fig:LRO_venn}
\end{center}

\section{Probabilistic Line Search}

In this section, I will briefly describe a hybrid inter/intra-step method for stochastic optimisation called \textbf{probabilistic line search} \cite{mahsereci2015probabilisticlinesearch}, proposed by Mahsereci et al in 2015. This method combines the structure of existing deterministic line searches with notions from Bayesian optimisation. Its hybrid nature comes from the fact that it searches within a single step of $g$ but over multiple iterations of $\xi$. I will explain the main idea, the theory behind it, and present an algorithm for its implementation.

\subsection{Theory}

To address the machine-learning motivation behind probabilistic line search, let us adopt some new notation: suppose $f(x, \xi)$ is a stochastic approximation of true objective function $F(x)$, with $F$ being too computationally expensive to optimise over directly. We can think of $f(x, \xi)$ as a \textit{mini-batch} loss function for which $F$ is \textit{full-batch}.

\subsubsection{Classical Line Search}

The idea behind \textit{classical line search} is to find an optimal step size $\gamma_k$ along a given search direction $g_k$ by evaluating $F(x_k + \hat{\gamma_k} g_k)$ across varying $\hat{\gamma_k}$. This is of course under the assumption that computation of $F(x)$ is tractable.

An inexact line search aims to satisfy certain conditions (such as sufficient \textit{decrease} or \textit{curvature}) that guarantee convergence and robustness. A popular choice of such termination conditions are \textbf{Wolfe conditions}, which deem $\hat{\gamma_k}$ as acceptable if it satisfies:
\begin{align}
    & F(x_k + \hat{\gamma_k} g_k) \leq F(x_k) + c_1 \hat{\gamma_k} g_k^T \nabla F(x_k), \label{eq:wolfe1} \\
    & g_k^T \nabla F(x_k + \hat{\gamma_k} g_k) \geq c_2 g_k^T \nabla F(x_k), \label{eq:wolfe2}
\end{align}
parameterised by scalars $c_1$ and $c_2$. When $g_k = \nabla F(x_k)$, such as in full-batch gradient descent, we can simplify the conditions to:
\begin{align}
    & F(x_k + \hat{\gamma_k} g_k) \leq F(x_k) + c_1 \hat{\gamma_k} \|g_k\|^2, \\
    & g_k^T \hat{g}_{k+1} \geq c_2 \|g_k\|^2.
\end{align}
Intuitively, one can think of the first (Armijo) condition as specifying that "\textit{$\hat{\gamma}$ should only increase if it improves $F$ faster than the line defined by $c_1$}"; and the second (curvature) condition as "\textit{$\hat{\gamma}$ should identify a basin at least as flat as $c_2$}".

With these conditions in place, the idea is to increment $\hat{\gamma_k}$ in exponentially-increasing steps until locating an opposing slope direction (negative $\hat{g}_{k+1}$). At this point, the algorithm narrows its search space and interpolates between previously-tried $\hat{\gamma_k}$ through, say, splines, until reaching a satisfying termination point. This process is described more formally in Algorithm \ref{alg:line_search}.

However, applying line searches directly to \textit{stochastic} optimisation is not straightforward. First, evaluating $F(x)$ exactly may be impractical or impossible due to noise or computational cost; second, uncertain gradients do not allow for a strict sequence of decisions collapsing the search space as line search intends.

\subsubsection{Introducing Probabilistics}

To overcome the challenges of stochastic line searches, the \textbf{probabilistic line search} (PLS) method combines the structure of the deterministic approach with notions from Bayesian optimisation. The crux of the algorithm is to update a Gaussian process ($\mathcal{GP}$) surrogate of the univariate optimisation objective:
\begin{equation}
    \hat{F}(\hat{\gamma_k}) = F(x_k + \hat{\gamma_k} g_k),
\end{equation}
where $g_k$ is a fixed search direction, using stochastic observations $\hat{f}_{j}(\hat{\gamma_k}) = f(x_k + \hat{\gamma_k} g_k, \xi_j)$ and $\hat{f}'_{j}(\hat{\gamma_k}) \in \frac{\partial \hat{f}_{j}(\hat{\gamma_k})}{\partial \hat{\gamma_k}}$.\footnote{The $j$ index can be thought of as selecting which mini-batch to compute loss over. Note that, with this notation, $\hat{f}_{j}(0) = f(x_k, \xi_j).$} This enables the use of a probabilistic belief over Wolfe conditions to monitor the descent, taking an exploration-exploitation trade-off between trying new values of $\hat{\gamma_k}$ or taking an existing one.

We can derive \textit{probabilistic} Wolfe conditions by first rephrasing eq \ref{eq:wolfe1} and \ref{eq:wolfe2} into quantities relative to 0:
\begin{align}
    & a_{\hat{\gamma_k}} = F(x_k) + c_1 \hat{\gamma_k} g_k^T \nabla F(x_k) - F(x_k + \hat{\gamma_k} g_k) \geq 0, \label{eq:0wolfe1} \\
    & b_{\hat{\gamma_k}} = g_k^T \nabla F(x_k + \hat{\gamma_k} g_k) - c_2 g_k^T \nabla F(x_k) \geq 0, \label{eq:0wolfe2}
\end{align}
and noting that these are linear projections of $F$ and $\nabla F$:
\begin{equation}
    \left[\begin{array}{l}
        a_{\hat{\gamma_k}} \\
        b_{\hat{\gamma_k}}
    \end{array}\right]=\left[\begin{array}{cccc}
        1 & c_1 \hat{\gamma_k} & -1 & 0 \\
        0 & -c_2 & 0 & 1
    \end{array}\right]\left[\begin{array}{c}
        F(x_k) \\
        g_k^T \nabla F(x_k) \\
        F(x_k + \hat{\gamma_k} g_k) \\
        g_k^T \nabla F(x_k + \hat{\gamma_k} g_k)
    \end{array}\right] \geq 0.
\end{equation}
Now, due to the linear properties of Gaussian processes, we can use the PLS-computed $\mathcal{GP}$ approximating $p(\hat{F}; \nabla \hat{F})$ to derive an expression for $p(a_{\hat{\gamma_k}}, b_{\hat{\gamma_k}})$. This enables computation of the final probabilistic Wolfe condition as a quadrant probability:\footnote{In practice, this probability is found by an integral over fixed bounds, approximating the quantity:
\begin{equation}
    p_{\hat{\gamma_k}}^{\text {Wolfe }} =
    \int_{-\frac{m_a^a}{\sqrt{C_{\hat{\gamma_k}}^a}}}^{\infty} \int_{-\frac{m^b}{\sqrt{C_{\hat{\gamma_k}}^{b b}}}}^{\infty} \mathcal{N}\left(\left[\begin{array}{l}
    a \\
    b
    \end{array}\right] ;\left[\begin{array}{l}
    0 \\
    0
    \end{array}\right],\left[\begin{array}{cc}
    1 & \rho_{\hat{\gamma_k}} \\
    \rho_{\hat{\gamma_k}} & 1
    \end{array}\right]\right) da \text{ } db,
\end{equation}
where $m$, $C$, and $\rho$ are metrics derived from the $\mathcal{GP}$. I will omit this full derivation for the sake of brevity and to avoid simply recreating proofs from the paper.}
\begin{equation}
    p_{\hat{\gamma_k}}^{\text {Wolfe }} =
    p(a_{\hat{\gamma_k}} > 0 \land b_{\hat{\gamma_k}} > 0).
\end{equation}

Finally, we can restate the Wolfe condition in terms of the probability $p_{\hat{\gamma_k}}^{\text {Wolfe }}$ and a new constant $c_W$ (the "Wolfe threshold"), such that an acceptable $\hat{\gamma_k}$ must be one that satisfies:
\begin{equation}
    p_{\hat{\gamma_k}}^{\text {Wolfe }} > c_w.
\end{equation}

\subsection{Algorithm}

With the key motivation of this method introduced, I use pseudocode to summarise the \textit{inexact} and \textit{probabilistic} line searches in Algorithms \ref{alg:line_search} and \ref{alg:prob_line_search} respectively. This format makes use of a number of items that should be clarified first:
\begin{itemize}
    \item $y_0$ and $y'_0$ represent the final $y$ and $y'$ of the optimiser's previous line search;
    \item $\sigma_{0}$ and $\sigma'_{0}$ represent the observed variance of $f$ and $f'$ thus far;
    \item \textfn{GetCandidates} identifies a list of potential candidates for $\hat{\gamma_k}$ by dividing the $\mathcal{GP}$ into $|\hat{\Gamma}|$ segments (each element of $\hat{\Gamma}$ being a boundary, plus one extrapolation boundary), taking the minima in each.
\end{itemize}

To make the probabilistic line search alterations clear, I highlight changes according to whether they concern \colorbox{LRcolour}{\textbf{evaluating candidates $\hat{\gamma_k}$}} (termination \& selection) or more specifically \colorbox{Dcolour}{\textbf{computing $\mathcal{GP}$}}.

\begin{minipage}[t]{.48\linewidth}
\begin{algorithm}[H]
\caption{Inexact Line Search}
\KwIn{$\hat{f}_{j}, y_0, y'_0, \gamma_{k-1}.$}
\vspace{0.25cm}
$\hat{\Gamma}, Y, Y' \gets$ \textfn{InitLists} $(0, y_0, y'_0)$\;
$\hat{\gamma_k} \gets \gamma_{k-1}$\;
\vspace{0.25cm}
\While{\textup{len}$(\hat{\Gamma}) < 10 $ \textup{\textbf{and} Wolfe conditions unmet}}{
    $y, y' \gets \hat{f}_{j}(\hat{\gamma_k}), \hat{f}'_{j}(\hat{\gamma_k})$\;
    $j \gets j+1$\;
    $\hat{\Gamma}, Y, Y' \gets$ \textfn{Append} $(\hat{\gamma_k}, y, y')$\;
    $p^{\text{Wolfe}} \gets \textfn{BoolWolfe}(y_0, y'_0, y, y')$\;
    \uIf{$p^{\textup{Wolfe}}$}{
        \Return $\gamma_{k} = \hat{\gamma_k}$\;
    }
    \Else{
        \uIf{$y' < 0$ \textup{and extrapolating}}{
            $\hat{\gamma_k} \gets 2\hat{\gamma_k}$\;
        }
        \Else{
          $\hat{\gamma_k} \gets \textfn{CubicMin}(\hat{\Gamma}, Y, Y')$\;
        }
    }
}

\Return $j$, and $\gamma_k, y_k, y'_k \in \hat{\Gamma}, Y, Y'$ that has minimal $y_k$\;
\vspace{0.25cm}
\KwResult{Effective $\gamma_k$ for step $k$.}
\label{alg:line_search}
\end{algorithm}
\end{minipage}
\hfill
\begin{minipage}[t]{.48\linewidth}
\begin{algorithm}[H]
\caption{Probabilistic Line Search}
\KwIn{$\hat{f}_{j}, y_0, y'_0, \hat{\gamma_{k-1}} \highlight[Dcolour]{\sigma_0, \sigma'_0}.$}
\vspace{0.25cm}
$\highlight[Dcolour]{\mathcal{GP} \gets \textfn{InitGP} (y_0, y'_0, \sigma_0, \sigma'_0)}$\;
$\hat{\Gamma}, Y, Y' \gets$ \textfn{InitLists} $(0, y_0, y'_0)$\;
$\hat{\gamma_k} \gets \hat{\gamma_{k-1}}$\;
\vspace{0.25cm}
\While{\textup{len}$(\hat{\Gamma}) < 10 $ \textup{\textbf{and} Wolfe conditions unmet}}{
    $y, y' \gets \hat{f}_{j}(\hat{\gamma_k}), \hat{f}'_{j}(\hat{\gamma_k})$\;
    $j \gets j+1$\;
    $\hat{\Gamma}, Y, Y' \gets$ \textfn{Append} $(\hat{\gamma_k}, y, y')$\;
    $\highlight[Dcolour]{\mathcal{GP} \gets \textfn{PosteriorGP}(\hat{\gamma_k}, y, y')}$\;
    $p^{\text{Wolfe}} \gets \highlight[LRcolour]{\textfn{ProbWolfe}(\hat{\Gamma}, \mathcal{GP})}$\;
    \uIf{$p^{\textup{Wolfe}}\highlight[LRcolour]{>c_w}$}{
        \Return $\gamma_{k} = \hat{\gamma_k}$\;
    }
    \Else{
        $\highlight[LRcolour]{\hat{\Gamma}_{\text{cand}} \gets \textfn{GetCandidates}(\mathcal{GP})}$\;
        $\highlight[LRcolour]{EI \gets \textfn{ExpectedImprovement}(\hat{\Gamma}_{\text{cand}}, \mathcal{GP})}$\;
        \vspace{-0.5cm}
        $\highlight[LRcolour]{PW \gets \textfn{ProbWolfe}(\hat{\Gamma}_{\text{cand}}, \mathcal{GP})}$\;
        \vspace{-0.5cm}
        $\highlight[LRcolour]{\hat{\gamma_k} \gets \argmax_{\hat{\Gamma}_{\text{cand}}} (PW \odot EI)}$\;
    }
}

\Return $j$, and $\gamma_k, y_k, y'_k \in \hat{\Gamma}, Y, Y'$ that has minimal $y_k$, \colorbox{Dcolour}{and the new $\sigma, \sigma'$}\;
\vspace{0.25cm}
\KwResult{Effective $\gamma_k$ for step $k$.}
\label{alg:prob_line_search}
\end{algorithm}
\end{minipage}

\section{D-Adaptation}

D-Adaptation is a novel \textit{inter-step learning rate optimisation} approach proposed by Defazio and Mishchenko in 2023 \cite{defazio2023dadaptation}. It is intended to be integrated as an extension on top of existing gradient-based optimisers like SGD, Adam, and AdaGrad. The core idea of D-Adaptation is to build up an estimate of the distance $D = \|x_0 - x_*\|$ over the course of a run, and compensate for that initial distance via dynamic adjustments to $\gamma_k$.

Unlike typical adaptive optimisers such as Adam and AdaGrad, D-Adaptation brings the significant advantage of not expecting the programmer to identify a fortuitous initial learning rate $\gamma_0$ beforehand, the idea being that this is compensated for through the discovery of $D$. However, unlike a line search, D-Adaptation tends towards an optimal $\gamma_k$ (under particular assumptions) on-the-fly, without requiring multiple evaluations of $f$ at each iterate.

\subsection{Theory}

As mentioned in \ref{sec:convex_opt}, providing theoretical guarantees for non-convex optimisation is challenging, but convex-designed optimisers like GD often translate well to non-convex scenarios empirically with minor adjustments. Defazio et al follow in this vein by deriving D-Adaptation and its guarantees in the theoretically-amenable context of $f$ being \textbf{convex} (eq \ref{eq:f_convex}), \textbf{deterministic} (eq \ref{eq:nonstoch_minima}) and \textbf{Lipschitz-continuous} (eq \ref{eq:lipschitz}); then later show that D-Adaptation applies well to more unpredictable functions empirically.

\begin{aside}
\asidetitle{Aside: Lipschitz functions} \label{sec:lipschitz}

A function $f : X \longmapsto Y$ is \textbf{Lipschitz Continuous} iff there exists some constant ("\textbf{Lipschitz constant}") $G$ such that:
\begin{equation} \label{eq:lipschitz}
    \abs{f(x_1) - f(x_2)} \leq G \abs{x_1 - x_2} \text{\quad for all } x_1, x_2 \in X,
\end{equation}
and this function can be called \textbf{bi-Lipschitz continuous} if its inverse $f^{-1}$ is also Lipschitz continuous. In other words, the rate-of-increase of Lipschitz functions is bounded by some constant $G$.

In many cases, we will be unable to derive an exact Lipschitz constant (for fully-connected neural networks, this is NP-hard \cite{spectral_normalisation}), but rather a \textbf{Lipschitz bound} $G \leq B$ which may be either global (independent on $x$) or local (dependent on $x$).

\end{aside}

Avoiding simply recreating the proofs from the D-Adaptation paper, I will focus on providing intuition behind their core findings.

\subsubsection{Subgradient Method Convergence}

Under our assumptions of convexity and Lipschitz-continuity, it is well known that the subgradient method (GD with eq \ref{eq:subgrad-method}) has the following convergence rate for constant step-size $\gamma$ \cite{subgradmethod-convergence, nesterov2018lectures}:
\begin{equation}
    f(x_k) - f(x_*) \leq \frac{D^2 + G^2 \gamma^2 k}{2 \gamma k},
\end{equation}
which tends to $G^2 \gamma/2$ as $k \rightarrow \infty$. Intuitively, this encapsulates the fact that:
\begin{itemize}
    \item Convergence will be slower when we initialise $x_0$ further away from $x_*$ (by $D^2$);
    \item A high non-diminishing learning rate will cause oscillations (by $G^2 \gamma^2 k$ and the eventuality $G^2 \gamma/2$); 
    \item High learning rates can improve convergence in the meantime (by denominator $2\gamma k$).
\end{itemize}
The more promising bound comes when we instead assume a diminishing learning rate $\gamma_k$, at which the convergence becomes:
\begin{equation}
    f(x_k) - f(x_*) \leq \frac{D^2 + G^2 \sum^{k}_{i=0} \gamma^2_i}{2 \sum^{k}_{i=0} \gamma_i},
\end{equation}
which tends to $0$ as $k \rightarrow \infty$. Excitingly, one can see that if we take this diminishing learning rate to be a simple fraction,
\begin{equation} \label{eq:optimal_lr}
    \gamma_k = \frac{D}{G \sqrt{k}},
\end{equation}
the bound now simplifies to just:
\begin{equation}
    f(x_k) - f(x_*) \leq \frac{D^2 + G^2 \sum^{k}_{i=0} \gamma^2_i}{2 \sum^{k}_{i=0} \gamma_i} = \frac{DG}{\sqrt{n}}.
\end{equation}
Thus we have convergence rate $\mathcal{O}(1/\sqrt{k})$ for the subgradient method when given eq \ref{eq:optimal_lr}'s learning rate, meaning it can achieve error $f(x_k) - f(x_*) \leq \epsilon$ in $\mathcal{O}(1/\epsilon^2)$ steps. This is the tightest worse-case bound achievable for the algorithm's complexity class \cite{nesterov2018lectures}.

\subsubsection{D-Adaptation Step Size}
Now acknowledging $\gamma_k = D / G \sqrt{k}$ (eq \ref{eq:optimal_lr}) as the ideal learning rate under our assumptions, we are faced with the problem of $D$ and $G$ being unknown quantities for general-case $f$.

A solution employed by AdaGrad \cite{duchi2011adagrad} is to effectively (under-)estimate Lipschitz constant $G$ by taking the root mean square $G \approx \sqrt{\sum^k_{i=0} \|g_i\|^2 / k}$, which when substituted into eq \ref{eq:optimal_lr} gives us the AdaGrad-Norm update:
\begin{equation} \label{eq:adagrad_lr}
    \gamma_k = \frac{D}{\sqrt{(\sum^k_{i=0} \|g_i\|^2)}}.
\end{equation}
However, this still comes with the expectation on the programmer to identify $D$ themselves through, for example, a grid search.

\begin{aside}
\asidetitle{Aside: Normalisation notation}

Defazio et al make use of an overloaded normalisation notation, referring to the weighted "A-matrix" norm when subscripted with a matrix:
\begin{equation}
    \left\|\mathbf{x}\right\|_A := \sqrt{\sum_{i=1}^n x_i^2 A_i},
\end{equation}
p-norm normalisation when subscripted with a scalar:
\begin{equation}
    \|\mathbf{x}\|_p:=\left(\sum_{i=1}^n\left|x_i\right|^p\right)^{1 / p},
\end{equation}
of course defaulting to the Euclidean norm when no subscript is given:
\begin{equation}
     \|\mathbf{x}\| := \|\mathbf{x}\|_2 = \sqrt{\sum_{i=1}^n x_i^2}.
\end{equation}
\end{aside}

D-Adaptation's core innovation is to build up an estimate of $D$ by constructing a \textbf{lower bound $D \geq \hat{d}_k$} at each step, taking the current best estimate to be $d_k = \max_i (\hat{d}_i)$. This lower bound is very informative: intuitively, the further away one knows $x_*$ to be from $x_0$, the more liberally step-size $\gamma_k$ can be set, particularly for early $k$.

Using our (under-)estimate $d_k$ of $D$, the D-Adaptation paper now takes the step-size to be:
\begin{equation} \label{eq:dadapt_sgd_lr}
    \gamma_k = \frac{d_k \phi_k}{\|g_0\|},
\end{equation}
where $\phi_k$ denotes a user-defined scale which can be left at 1.0 in the majority of cases, or optionally scheduled (e.g. with a learning rate scheduling technique in Figure \ref{fig:LRO_venn}). The choice of Lipschitz proxy $\|g_0\|$ is less robust to noise than AdaGrad's normaliser, but more likely to be close to $G$ in convex scenarios (as the first gradient should be the highest for reasonable step size).

\begin{aside}
    \asidetitle{Aside: Learning rate notation}

    For readers referring to the D-Adaptation paper, it is important to note that at this point my notation for learning rate is diverging from the authors' in a slightly subtle manner.

    Throughout the latter half of the paper, the authors use $\lambda_k$ to denote step-size instead of $\gamma_k$, and instead use $\gamma_k$ to denote a \textbf{scaling factor} that I call $\phi_k$. I found this alteration to be confusing, so instead leave $\gamma_k$ as the step size for the purposes of this report.
\end{aside}

\subsubsection{Core Equation}

For stochastic gradient descent,\footnote{Note that the paper first derives this bound in the more theoretically-sound context of \textit{dual averaging}, reaching the following formulation:
\begin{equation}
    D \geq \hat{d}_{k+1}=\frac{\phi_{k+1}\left\| \sum_{i=0}^k \gamma_i g_i \right\|^2-\sum_{i=0}^k \phi_i d_i^2\left\|g_i\right\|^2}{2\left\| 
 \sum_{i=0}^k \gamma_i g_i \right\|}.
\end{equation}
I will instead focus on addressing the SGD-optimized formulation (eq \ref{eq:SGD-dk}), which is more useful in practice as well as more interpretable.
} the core D-Adaptation equation is their derivation for $\hat{d}_{k+1}$:
\begin{equation} \label{eq:SGD-dk}
    \hat{d}_{k+1}=\frac{\left\| \sum_{i=0}^k \gamma_i g_i \right\|^2 - \sum_{i=0}^k \phi_i^2 d_i^2\left\|g_i\right\|^2 / \; \|g_0\|^2}{\left\| 
 \sum_{i=0}^k \gamma_i g_i \right\|},
\end{equation}
where each $\gamma_i$ is arbitrary but positive. While this result may look a little inscrutable, one can achieve an intuition for it after breaking down its terms in context.

\definecolor{displacement}{RGB}{200,215,245}
\definecolor{displacement_norm}{RGB}{215,210,250}
\definecolor{effort}{RGB}{210,235,240}

To begin with, when taking $\gamma_i=\frac{d_i \phi_i}{\|g_0\|}$ per eq \ref{eq:dadapt_sgd_lr}, this simplifies to:
\begin{equation}
    \hat{d}_{k+1}=\frac
        {\highlight[displacement]{\left\| \sum_{i=0}^k \gamma_i g_i \right\|^2} - \highlight[effort]{\sum_{i=0}^k \gamma_i^2 \left\|g_i\right\|^2}}
        {\highlight[displacement_norm]{\left\| \sum_{i=0}^k \gamma_i g_i \right\|}}.
\end{equation}
We can examine each term informally:
\begin{itemize}
    \item On the numerator, $\highlight[displacement]{\left\| \sum_{i=0}^k \gamma_i g_i \right\|^2}$ is a measurement of the \textbf{displacement} of $x_{k+1}$ from $x_0$.
    \item Conversely, $\highlight[effort]{\sum_{i=0}^k \gamma_i^2 \left\|g_i\right\|^2}$ measures the \textbf{difficulty} of reaching $x_{k+1}$: routes that are more circuitous (with more positive/negative fluctuation in $\gamma_k g_i$, or higher variability in $\gamma_k g_i$ as picked up on by $\gamma_i^2 \left\|g_i\right\|^2$) will have increased \textit{difficulty} but decreased \textit{displacement}.
    \item By subtracting difficulty from displacement, we can see the numerator as a measurement of how \textbf{straight-forward} the path from $x_0$ to $x_{k+1}$ has been, normalised across path lengths by the denominator $\highlight[displacement_norm]{\left\| \sum_{i=0}^k \gamma_i g_i \right\|}$.
\end{itemize}

The idea here is that, for convex or near-convex $f$, we expect the optimisation path to be more \textit{straight-forward} when $x_0$ is further away from $x_*$, as there is a lower risk of experiencing oscillations. Intuitively, as the D-Adaptation optimiser explores $f$, it discovers the \textit{straight-forwardness} of the region near $x_0$, thus indicating a lower bound for $D$.

\subsection{Algorithm}
Finally, we can take a look at how D-Adaptation is applied in practice. In Algorithm \ref{alg:sgd_dadapt}, I show how this extension can be integrated with stochastic gradient descent. To make the D-Adaptation extensions clear, I highlight changes according to whether they concern \colorbox{LRcolour}{\textbf{formulation of $x$'s learning rate}} or more specifically \colorbox{Dcolour}{\textbf{computing $d_{k+1}$}}.

\begin{minipage}[t]{.46\linewidth}
\begin{algorithm}[H]
\caption{SGD}
\KwIn{$x_0$, \\$\gamma_k$ (default $10^{-3}$).}
\For{$k\gets0$ \KwTo $n$}{
    $g_k \in \partial f\left(x_k, \xi_k\right)$\;
    $x_{k+1}\gets x_k-\gamma_k g_k$\;
}
\KwResult{$x_n$ that estimates $x_*$.}
\label{alg:sgd}
\end{algorithm}
\end{minipage}
\hfill
 \begin{minipage}[t]{.46\linewidth}
 \begin{algorithm} [H]
\caption{SGD w/ D-Adaptation}
\KwIn{$x_0$, \\$\highlight[LRcolour]{\phi_k \; (\text{default} \; 1.0)}$, \\$\highlight[Dcolour]{d_0 \; (\text{default} \; 10^{-6})}$.}
\For{$k\gets0$ \KwTo $n$}{
    $g_k \in \partial f\left(x_k, \xi_k\right)$\;
    $\highlight[LRcolour]{\gamma_k\gets\frac
        {d_k \phi_k}
        {\left\|g_0\right\|}}$\;
    $x_{k+1}\gets x_k - \gamma_k g_k$\;
    \colorbox{Dcolour}{\textit{\underline{Updating D-bounds:}}} \\
    $\highlight[Dcolour]{s_{k+1}\gets s_k + \gamma_k g_k}$\;
    $\highlight[Dcolour]{r_{k+1}\gets r_k + \gamma_k^2 \left\|g_k\right\|^2}$\;
    $\highlight[Dcolour]{\hat{d}_{k+1}\gets\frac
        {\left\|s_{k+1}\right\|^2-r_{k+1}}
        {\left\|s_{k+1}\right\|}}$\;
    $\highlight[Dcolour]{d_{k+1}\gets\max \left(d_k, \hat{d}_{k+1}\right)}$\;
}
\KwResult{$x_n$ that estimates $x_*$.}
\label{alg:sgd_dadapt}
\end{algorithm}
\end{minipage}

\section{Discussion} \label{sec:discussion}

By unifying their notation and presentation, I have aimed to highlight a correspondence between the way in which both \textit{PLS} (Algorithm \ref{alg:prob_line_search}) and \textit{D-Adaptation} (Algorithm \ref{alg:sgd_dadapt}) will \\\colorbox{Dcolour}{accumulate knowledge of $f$ into $\mathcal{GP}$ or $d_k$}, and then \colorbox{LRcolour}{apply the item to find viable $\gamma_k$}.

This approach is typical of probabilistic numerics: we take a classical algorithm; find an efficient way to extract variance over its search space; use this variance to establish a $\mathcal{GP}$ posterior; and use expected improvement on the posterior to find points that reduce uncertainty.

A similar pattern emerges in D-Adaptation's methodology: they take a classical algorithm; find an efficient way to extract \textit{bounds} over its search space; and use these bounds to explore points closer to the optimum.

\subsection{Combining Approaches} 

We can now consider means of combining the two approaches. At the most basic level, Algorithm \ref{alg:sgd_pls_dadapt} takes the D-Adaptation algorithm (defined in \ref{alg:sgd_dadapt}) and initialises a probabilistic line search at each iteration, starting from the D-Adapted $\gamma_k$ value.

There is a big assumption here that D-Adaptation's notion of \textit{difficulty} and its relation to $D$ will still transfer when the step-size is extended by PLS; in reality, one may wish to alter the $r_{k+1}$ definition to include metrics from the PLS search, such as the number of PLS-iterations $j$ taken to reach a satisfying point. I use colour to highlight the changes made to \colorbox{MixColour}{\textbf{integrate PLS into D-Adaptation}}.

We can also consider how to exploit D-Adaptation's $D$ bounds within PLS. While there are many approaches one could take, with varying degrees of confidence, one particularly simple integration would be to use $d_k$ to limit the extended segment boundary used by \textfn{GetCandidates}, preventing the search from venturing too far back to $x_0$.

In other words, where \textfn{GetCandidates} finds the $\mathcal{GP}$ minima within segments defined by the boundaries:
\begin{equation}
    \hat{\Gamma} \cup \textfn{Extend}(\max(\hat{\Gamma})),
\end{equation}
where \textfn{Extend} is a simple operation such as scalar multiplication, our new \textfn{D-GetCandidates} would use $d_k$ to limit the extension:
\begin{equation}
    \hat{\Gamma} \cup \max(x_0 - d_k g_k, \textfn{Extend}(\max(\hat{\Gamma}))),
\end{equation}
with max taking the maximum by the norms of its two arguments.

I again use colour to highlight the changes made to \colorbox{MixColour}{\textbf{integrate D-Adaptation into PLS}}, detailed in Algorithm \ref{alg:dinformed_pls}.

 \begin{minipage}[t]{.46\linewidth}
 \begin{algorithm} [H]
\caption{SGD-PLS w/ D-Adapt}
\KwIn{$x_0$, \\$\highlight[LRcolour]{\phi_k \; (\text{default} \; 1.0)}$, \\$\highlight[Dcolour]{d_0 \; (\text{default} \; 10^{-6})}$.}
$\highlight[MixColour]{\sigma_0, \sigma'_0 \gets \textfn{EstimateVar}(x_0)}$\;
$\highlight[MixColour]{j \gets 0}$\;
\For{$k\gets0$ \KwTo $n$}{
    $g_k \in \partial f\left(x_k, \highlight[MixColour]{\xi_j}\right)$\;
    $\highlight[LRcolour]{\gamma_k\gets\frac
        {d_k \phi_k}
        {\left\|g_0\right\|}}$\;
    $\highlight[MixColour]{\hat{f}_{j} \gets [\lambda (\gamma) : f(x_k + \gamma g_k, \xi_j)]}$\;
    $\highlight[MixColour]{j, \gamma_k, y_k, y'_k, \sigma_{k+1}, \sigma'_{k+1}}$ $\highlight[MixColour]{\gets \textfn{PLS}(\hat{f}_{j}, y_k, y'_k, \gamma_k, \sigma_{k}, \sigma'_{k}, d_k)}$\;
    $x_{k+1}\gets x_k - \gamma_k g_k$\;
    \colorbox{Dcolour}{\textit{\underline{Updating D-bounds:}}} \\
    $\highlight[Dcolour]{s_{k+1}\gets s_k + \gamma_k g_k}$\;
    $\highlight[Dcolour]{r_{k+1}\gets r_k + \gamma_k^2 \left\|g_k\right\|^2}$\;
    $\highlight[Dcolour]{\hat{d}_{k+1}\gets\frac
        {\left\|s_{k+1}\right\|^2-r_{k+1}}
        {\left\|s_{k+1}\right\|}}$\;
    $\highlight[Dcolour]{d_{k+1}\gets\max \left(d_k, \hat{d}_{k+1}\right)}$\;
}
\KwResult{$x_n$ that estimates $x_*$.}
\label{alg:sgd_pls_dadapt}
\end{algorithm}
\end{minipage}
\hfill
\begin{minipage}[t]{.48\linewidth}
\begin{algorithm}[H]
\caption{D-Informed PLS}
\KwIn{$\hat{f}_{j}, y_0, y'_0, \hat{\gamma_{k-1}} \highlight[Dcolour]{\sigma_0, \sigma'_0}, \highlight[MixColour]{d_k}.$}
\vspace{0.25cm}
$\highlight[Dcolour]{\mathcal{GP} \gets \textfn{InitGP} (y_0, y'_0, \sigma_0, \sigma'_0)}$\;
$\hat{\Gamma}, Y, Y' \gets$ \textfn{InitLists} $(0, y_0, y'_0)$\;
$\hat{\gamma_k} \gets \hat{\gamma_{k-1}}$\;
\vspace{0.25cm}
\While{\textup{len}$(\hat{\Gamma}) < 10 $ \textup{\textbf{and} Wolfe conditions unmet}}{
    $y, y' \gets \hat{f}_{j}(\hat{\gamma_k}), \hat{f}'_{j}(\hat{\gamma_k})$\;
    $j \gets j+1$\;
    $\hat{\Gamma}, Y, Y' \gets$ \textfn{Append} $(\hat{\gamma_k}, y, y')$\;
    $\highlight[Dcolour]{\mathcal{GP} \gets \textfn{PosteriorGP}(\hat{\gamma_k}, y, y')}$\;
    $p^{\text{Wolfe}} \gets \highlight[LRcolour]{\textfn{ProbWolfe}(\hat{\Gamma}, \mathcal{GP})}$\;
    \uIf{$p^{\textup{Wolfe}}\highlight[LRcolour]{>c_w}$}{
        \Return $\gamma_{k} = \hat{\gamma_k}$\;
    }
    \Else{
        $\highlight[LRcolour]{\hat{\Gamma}_{\text{cand}} \gets} \highlight[MixColour]{\textfn{D-GetCandidates}(\mathcal{GP}, d_k)}$\;
        \vspace{-0.5cm}
        $\highlight[LRcolour]{EI \gets \textfn{ExpectedImprovement}(\hat{\Gamma}_{\text{cand}}, \mathcal{GP})}$\;
        \vspace{-0.5cm}
        $\highlight[LRcolour]{PW \gets \textfn{ProbWolfe}(\hat{\Gamma}_{\text{cand}}, \mathcal{GP})}$\;
        \vspace{-0.5cm}
        $\highlight[LRcolour]{\hat{\gamma_k} \gets \argmax_{\hat{\Gamma}_{\text{cand}}} (PW \odot EI)}$\;
    }
}

\Return $j$, and $\gamma_k, y_k, y'_k \in \hat{\Gamma}, Y, Y'$ that has minimal $y_k$, \colorbox{Dcolour}{and the new $\sigma, \sigma'$}\;
\vspace{0.25cm}
\KwResult{Effective $\gamma_k$ for step $k$.}
\label{alg:dinformed_pls}
\end{algorithm}
\end{minipage}

\subsection{Conclusion}

In summary, we have motivated the methods of \textit{D-Adaptation} and \textit{probabilistic line search} as learning-rate-free LRO approaches, and identified their common design patterns through a unified notation. In Algorithms \ref{alg:sgd_pls_dadapt} and \ref{alg:dinformed_pls}, I described a rudimentary proposal for an algorithm that can combine the two methods, with the intent that it might adopt the \textit{stochastic resiliance} of PLS in combination with the \textit{inter-step convergence} of D-Adapt to help steer the search.

While this combined algorithm would likely require additional work to be applicable in practice, I hope it might serve as an insightful case-study into how the probabilistic numerics methodology of \textit{exploiting data to reduce uncertainty} might be implemented and extended across contexts.

\newpage
\appendix
\bibliographystyle{plain}
\bibliography{reference}

\end{document}